\documentclass[sigconf]{acmart}

\usepackage{graphicx}
\usepackage{amsmath}
\usepackage{amssymb}
\usepackage{amsthm}
\usepackage{booktabs}
\usepackage{algorithm}
\usepackage{algorithmic}
\usepackage{enumerate}
\usepackage{bbm}
\usepackage{multirow}
\usepackage[FIGTOPCAP]{subfigure}
\usepackage{balance}
\def\BibTeX{{\rm B\kern-.05em{\sc i\kern-.025em b}\kern-.08emT\kern-.1667em\lower.7ex\hbox{E}\kern-.125emX}}
    
\setcopyright{acmcopyright}

\begin{document}
\fancyhead{}
%
\title{Alleviating Feature Confusion for \\Generative Zero-shot Learning}

\author{Jingjing Li}
\affiliation{\institution{University of Electronic Science and Technology of China}
\institution{The University of Queensland}}
\email{lijin117@yeah.net}

\author{Mengmeng Jing}
\affiliation{\institution{University of Electronic Science and Technology of China}}
\email{jingmeng1992@gmail.com}

\author{Ke Lu}
\affiliation{\institution{University of Electronic Science and Technology of China}}
\email{kel@uestc.edu.cn}

\author{Lei Zhu}
\affiliation{\institution{Shandong Normal University}}
\email{leizhu0608@gmail.com}

\author{Yang Yang}
\affiliation{\institution{University of Electronic Science and Technology of China}}
\email{dlyyang@gmail.com}

\author{Zi Huang}
\affiliation{\institution{The University of Queensland}}
\email{huang@itee.uq.edu.au}
%
\renewcommand{\shortauthors}{Jingjing Li et al.}
\renewcommand{\shorttitle}{Alleviating Feature Confusion for Generative ZSL}

%
\begin{abstract}
Lately, generative adversarial networks (GANs) have been successfully applied to zero-shot learning (ZSL) and achieved state-of-the-art performance. By synthesizing virtual unseen visual features, GAN-based methods convert the challenging ZSL task into a supervised learning problem. However, GAN-based ZSL methods have to train the generator on the seen categories and further apply it to unseen instances. An inevitable issue of such a paradigm is that the synthesized unseen features are prone to seen references and incapable to reflect the novelty and diversity of real unseen instances. In a nutshell, the synthesized features are confusing. One cannot tell unseen categories from seen ones using the synthesized features. As a result, the synthesized features are too subtle to be classified in generalized zero-shot learning (GZSL) which involves both seen and unseen categories at the test stage. In this paper, we first introduce the feature confusion issue. Then, we propose a new feature generating network, named alleviating feature confusion GAN (AFC-GAN), to challenge the issue. Specifically, we present a boundary loss which maximizes the decision boundary of seen categories and unseen ones. Furthermore, a novel metric named feature confusion score (FCS) is proposed to quantify the feature confusion. Extensive experiments on five widely used datasets verify that our method is able to outperform previous state-of-the-arts under both ZSL and GZSL protocols. 
\end{abstract}

%
%


\copyrightyear{2019} 
\acmYear{2019} 
\acmConference[MM '19]{Proceedings of the 27th ACM International Conference on Multimedia}{October 21--25, 2019}{Nice, France}
\acmBooktitle{Proceedings of the 27th ACM International Conference on Multimedia (MM '19), October 21--25, 2019, Nice, France}
\acmPrice{15.00}
\acmDOI{10.1145/3343031.3350901}
\acmISBN{978-1-4503-6889-6/19/10}


%

%

%
\maketitle

\section{Introduction}
Conventional machine learning paradigms assume that the testing categories are included in the training set. For instance, a machine learning algorithm is able to recognize a {horse} since some horse images are observed for training. However, this assumption does not always hold in real-world applications~\cite{lampert2014attribute,xian2017zero}. Real-world application is like a box of chocolates. You never know what you are going to get. The testing instances can come from either {\it seen} categories or {\it unseen} categories. A well-known learning scheme proposed to handle this situation is called zero-shot learning (ZSL)~\cite{zhang2016zero,socher2013zero,ding2018generative,kodirov2017semantic,xian2018feature}. ZSL methods challenge the unseen instances by leveraging side information, e.g., semantic attributes and text descriptions. Thus, a typical ZSL task involves a visual space and a semantic space, where only seen samples are available in the visual space and both seen and unseen samples share the semantic space~\cite{xian2017zero}. 

Existing ZSL methods can be roughly grouped into two categories: {\it embedding methods} and {\it generative methods}. Embedding methods~\cite{zhang2017learning,ding2018generative,zhang2016zero} generally learn a visual-to-semantic, semantic-to-visual or joint embedding space so that both the seen and unseen samples can be compared in the shared space. Generative methods~\cite{verma2018generalized,zhu2018generative,xian2018feature} aim to synthesize unseen features directly from the semantic descriptions. Recently, generative methods have achieved state-of-the-art results by converting ZSL into a general supervised learning problem. Notably, there are also several embedding methods~\cite{ding2018generative} deploying generative networks for data augmentation.

State-of-the-art generative methods are based on GANs~\cite{goodfellow2014generative,gulrajani2017improved}. Since there are no visual features for unseen categories, the GAN generator is mainly trained on seen categories. Specifically, at the training stage, the GAN generator synthesizes virtual features of seen categories from their corresponding semantic embeddings to fool the GAN discriminator. Considering that seen categories and unseen categories share the same semantic space, at the testing stage, it is expected that the trained generator is able to synthesize meaningful features for the unseen categories as well. However, the apple never falls far from the tree. With the GANs trained on the seen categories, the synthesized features of unseen categories are prone to seen references and incapable to reflect the novelty and diversity of real unseen instances. As a result, the synthesized features are too confusing to be classified in generalized zero-shot learning (GZSL)~\cite{chao2016empirical}, which means that the testing set consists of both seen and unseen instances instead of only unseen ones as in classical ZSL. For instance, state-of-the-art method GAZSL~\cite{zhu2018generative} leverages GANs to generate unseen features from noisy texts. GAZSL is able to achieve a ZSL accuracy of 68.2\% on AwA dataset~\cite{lampert2009learning}. However, when it comes to GZSL settings, the accuracy on unseen categories of GAZSL dramatically drops to 19.2\%. A tremendous 72\% (49\% of 68.2\%) performance drops only because seen categories are involved in. The superficial reason of performance dropping is that many unseen instances are misclassified into seen categories. However, the culprit of such a failure is that the synthesized unseen features are confusing with seen features. For convenience, we name such a phenomenon as {\it feature confusion} issue in generative ZSL.

In this paper, we propose a novel feature generating network to alleviate the feature confusion problem. Specifically, we introduce a boundary loss into the GANs to explicitly maximize the decision boundary of seen and unseen categories. It is worth noting that there is NO overlap between the semantic labels of seen categories and unseen categories. Thus, it does make sense to maximize the feature decision boundary between them. In addition, we also introduce a multi-modal consistent loss to promote the diversity and preserve the semantic consistency of synthesized features. Specifically, we encourage the generated features to be able to be translated back to their semantic embeddings. Since the semantic information of seen and unseen categories are different, it can alleviate the feature confusion by forcing the synthesized features back to their corresponding semantic embeddings, i.e., synthesized seen features back to seen semantics and unseen features back to unseen semantics. In summary, the main contributions of this paper are:
\vspace{-5pt}
\begin{enumerate}[1)]
\item We propose a novel generative method named {\it alleviating feature confusion GAN} (AFC-GAN) for zero-shot learning by taking advantage of generative adversarial networks. Compared with existing GAN-based methods, we introduce a boundary loss which estimates and then maximizes the decision boundary of seen and unseen features. In other words, we minimize the feature confusion between the seen and unseen categories.  
\item We introduce a multi-modal cycle-consistent loss in AFC-GAN to promote the diversity and preserve the semantic consistency of synthesized features. Compared with existing GAN-based methods, our approach encourages that the synthesized features can be translated back to original semantic embeddings by a deep nonlinear mapping. As a result, the mode collapse issue and the feature confusion issue can be further alleviated.
\item Since generative zero-shot learning is a fresh topic in the community, there are very few existing researches noticing the feature confusion issue. We manifest the issue and design a new metric named {\it feature confusion score} (FCS) to quantify the phenomenon, which is expected to benefit subsequent researches.   
\item Extensive experiments on five widely used standard benchmarks, in both classical ZSL and generalized ZSL (GZSL) settings, verify that our proposed AFC-GAN is able to outperform previous state-of-the-art approaches with significant advantages.
\end{enumerate}


\section{Related Work}
\subsection{Zero-shot Learning}
Zero-shot learning (ZSL)~\cite{xian2017zero,zhang2016zero,ding2018generative,kodirov2017semantic,li2019zero} is one of the most exciting topics in the computer vision and multimedia communities since it deals with {\it unseen} or {\it novel} instances. Specifically, ZSL handles an open set problem where some test categories are not included in the training set. In general, an additional semantic dataset is introduced as side information to facilitate the zero-shot recognition. Thus, ZSL tasks involve a shared semantic space and a visual space where seen samples and unseen ones have distinctive data distributions.

According to the working mechanisms, existing ZSL methods can be grouped into either {\it embedding methods}~\cite{akata2015evaluation,dinglow2017,zhang2016zero,yang2016zero} or {\it generative methods}~\cite{verma2018generalized,zhu2018generative,xian2018feature,Li_2019_CVPR}. Specifically, embedding methods learn a visual-to-semantic embedding space, or a semantic-to-visual embedding space, or a shared intermediate embedding space where the two domains are connected. Generative methods directly synthesize virtual features for the unseen categories. Currently, most state-of-the-art methods are GAN-based generative approaches. There is also a few efforts~\cite{ding2018generative} deploying generative methods for data augmentation and then performing ZSL recognition by the embedding methods. In addition, ZSL can also be regarded as a special case of transfer learning~\cite{li2018heterogeneous,li2019locality,li2016low}.

Classical ZSL protocol assumes that the information of seen and unseen is known for a testing sample. So, it only focuses on unseen samples at the test stage. However, this assumption does not always hold in real-world applications. A more common situation is that the test dataset is a mixture of both seen and unseen instances. To handle this, generalized zero-shot learning (GZSL) protocol is proposed as a more challenging extension of classical ZSL. Compared with ZSL, GZSL needs to handle wider semantic space and larger testing samples. Another challenge of GZSL is that the features of seen samples and unseen samples are confusing. The method has to balance the accuracy of seen and unseen categories.


\begin{figure*}[t!p]
\begin{center}
\includegraphics[width=0.88\linewidth]{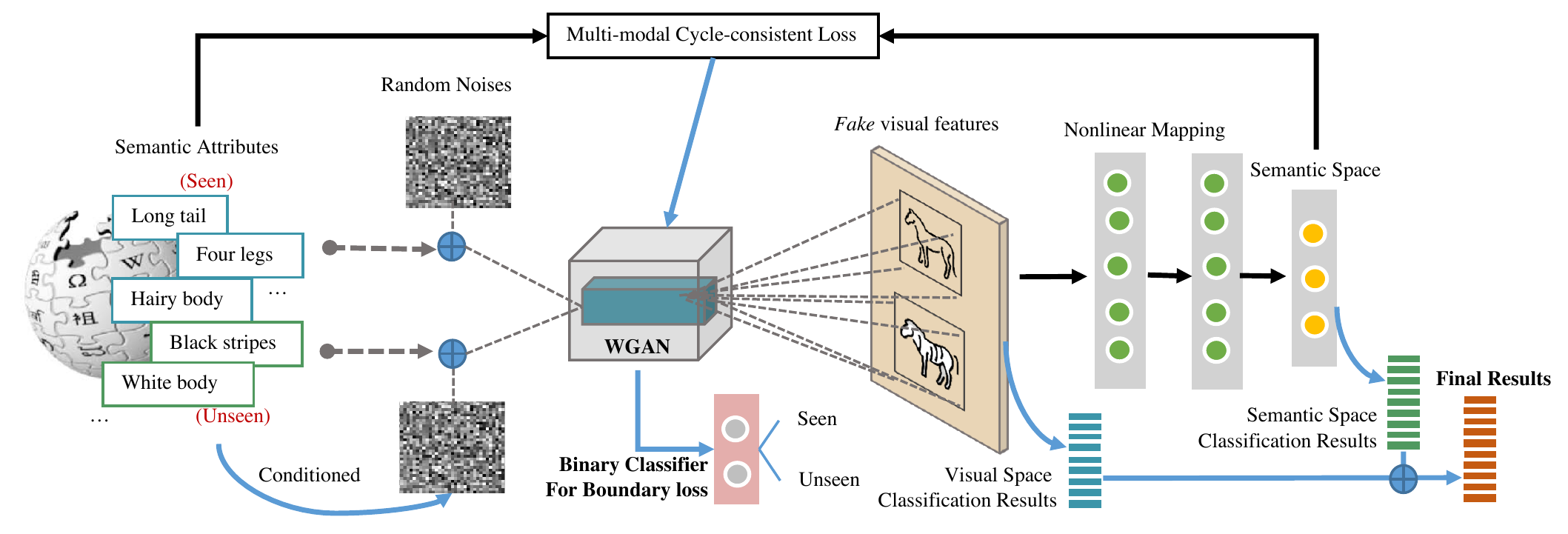}
\end{center}
\vspace{-10pt}
\caption{\small Ideas illustration of our alleviating feature confusion GAN (AFC-GAN). 1) The basis of our model is a conditional WGAN. 2) To alleviate the feature confusion issue, we introduce a boundary loss to maximize the decision boundary between seen and unseen categories. 3) To mitigate the mode collapse issue, we add a multi-modal consistency loss by a nonlinear mapping from visual space to semantic space. 4) We take full advantage of each component in our model and report the final results from both visual and semantic classifications.}
\label{fig:ideaill}
\end{figure*} 

\subsection{Generative Zero-shot Learning}
Generative zero-shot learning methods handle ZSL tasks by taking advantage of GANs. The main challenge of ZSL comes from the limitation that no unseen visual samples are available at the training stage. By leveraging GANs, one can synthesize unseen visual features from noises. For instance, Xian et al.~\cite{xian2018feature} propose a feature generating network for ZSL by deploying conditional WGAN. Zhu et al.~\cite{zhu2018generative} introduce a feature synthesizing network by GANs constrained by a visual pivot. Verma et al.~\cite{verma2018generalized} propose to handle GZSL by synthesized samples. It is worth noting that the mentioned methods are all published very recently. Generative zero-shot learning is a fresh yet very promising topic.

Although generative ZSL methods have achieved state-of-the-art performance, there are still two issues in previous work. The first issue is that the synthesized unseen features are prone to seen ones since the GAN is mainly trained on the seen samples. We name this issue as feature confusion. Feature confusion is not necessarily serious in classical ZSL where only unseen instances are involved for testing. However, feature confusion is a fatal issue in GZSL where the testing data consists of both seen and unseen instances. If the synthesized unseen features are prone to seen ones, unseen samples would be misclassified into seen categories. Many evidences can be found in the experimental section. For instance, most compared methods are able to achieve a satisfactory result in ZSL. However, the accuracy of unseen samples would dramatically drop in GZSL. As we mentioned before, the unseen samples accuracy of GAZSL~\cite{zhu2018generative} drops from 68.2\% (ZSL) to 19.2\% (GZSL). To the best of our knowledge, there is no previous work {\it explicitly} addressing the feature confusion issue in generative ZSL.  

The second issue in generative ZSL is the mode collapse problem which describes the phenomenon that synthesized samples have very low diversity. Mode collapse is a common problem in GANs. Previous work~\cite{isola2017image} show that this issue can be alleviated by image-to-image cycle-consistency. However, we do not have the unseen images at the training stage. Inspired by the very recent work~\cite{felix2018multi}, we introduce a multi-modal cycle-consistent loss into our GAN. Specifically, we encourage that the synthesized features can be translated back to their original semantic embeddings. Different from~\cite{felix2018multi} which only deploys a linear regressor for reconstruction, our method trains a nonlinear deep embedding model. Experiments verify that our method is significantly better than previous work.

\section{The Proposed Method}
\subsection{Notations and Definitions}
In zero-shot learning (ZSL), we have a visual space $\mathcal{X}$ and a semantic space $\mathcal{A}$. In this paper, we use $x$ to denote a visual feature and $a$ to denote a semantic embedding. The visual features are generally CNN features extracted from pre-trained deep networks, e.g., ResNet-50~\cite{he2016deep}. The semantic embeddings can be binary/numeric vectors or word embedding/RNN features. Both the visual samples and the semantic embeddings are further split into seen part $\{X, A\}$ and unseen part $\{X_u, A_u\}$. In zero-shot learning, we learn a function $f:(\mathcal{X},\mathcal{A}) \rightarrow \mathcal{Y}$, where $\mathcal{Y}$ is the label space. Then, in classical zero-shot learning, the function $f$ is applied to the unseen samples, i.e., $f:(\mathcal{X}_u,\mathcal{A}_u) \rightarrow \mathcal{Y}_u$. In generalized zero-shot learning, the function $f$ is applied to both seen and unseen samples, i.e., $f:(\mathcal{X}\cup\mathcal{X}_u,\mathcal{A}\cup\mathcal{A}_u) \rightarrow \mathcal{Y}\cup\mathcal{Y}_u$.

Generative zero-shot learning aims to train a GAN generator $G$ which is able to generate virtual visual samples $\hat{x}$ from random noise $z$ which is conditioned by semantic embedding $a$, i.e., $\hat{x}=G(z,a)$. Then, unseen visual samples can be directly handled by the classifier trained on synthesized unseen samples. It is worth noting that we synthesize virtual features rather image pixels. Previous work~\cite{xian2018feature} has verified that synthesizing features is more effective and efficient than synthesizing image pixels in real world applications.

\subsection{Overall Idea}
In this paper, we propose a novel generative ZSL method named {\it alleviating feature confusion GAN} (AFC-GAN). The core ideas of AFC-GAN consists of four parts: 1) We first deploy a conditional Wasserstein GAN (WGAN)~\cite{arjovsky2017wasserstein} to synthesize virtual features from random noises $z\sim \mathcal{N}(0,1)$ which is conditioned by the semantic embeddings $a \in \{A, A_u\}$. 2) To alleviate the feature confusion issue, we introduce a boundary loss to maximize the decision boundary of seen and unseen categories. Maximizing boundary loss is equivalent to minimizing feature confusion. 3) To mitigate the mode collapse issue, we further introduce a multi-modal cycle-consistent loss by encouraging that the synthesized features can be translated back to their corresponding semantic embeddings. Notably, the cycle loss can also alleviate feature confusion by forcing synthesized features to their corresponding categories. 4) Taking the above three ideas into consideration, we propose a novel ZSL method AFC-GAN as illustrated in Fig.~2. In the remainder of this section, we first report the basic architecture of our conditional WGAN. Then, we present the boundary loss and cycle-loss in details. At last, we formulate our overall objective function and propose a new metric named feature confusion score to quantify the feature confusion issue.

\subsection{The Discriminative WGAN}
The basic architecture of our model is a conditional WGAN. Specifically, we train a generator $G$ which synthesizes virtual visual features $\hat{x}$ from Gaussian noises $z\sim \mathcal{N}(0,1)$ and semantic embeddings. The semantic embedding is used as a class condition of the virtual features. At the same time, a discriminator $D$ is trained to distinguish the synthesized from the real. Since we do not have real unseen samples, the GAN is mainly trained on the seen categories. Recently, Xian et al.~\cite{xian2018feature} reported that a classification loss can help the generator $G$ to learn more discriminative features. In this paper, we also take the discriminative ability into account. As a result, we can train the discriminative WGAN by optimizing the following loss: 
\begin{equation}
\label{eq:dis}
  \begin{array}{l} 
 \min\limits_{G}\max\limits_{D} \mathcal{L}_{WGAN}=  \mathbb{E}[ D(x)]-\mathbb{E}[ D(G(z,a))] \\ \hspace{70pt} - \lambda (\mathbb{E}[\mathrm{log} P(y|G(z,a))] + \mathbb{E}[\mathrm{log} P(y|x)]) \\
 \hspace{70pt} - \beta \mathbb{E}[(\|\nabla_{\hat{x}} D(\hat{x})\|_2-1)^2],
  \end{array} 
\end{equation} 
where $\lambda>0$ and $\beta>0$ are two hyper-parameters, $G(z,a)$ indicates the synthesized features from $z$ which is conditioned by $a$. The part $\lambda (\mathbb{E}[\mathrm{log} P(y|G(z,a))]+ \mathbb{E}[\mathrm{log} P(y|x)])$ is the classification loss which encourages $G$ to generate discriminative features. The last term $\mathbb{E}[(\|\nabla_{\hat{x}} D(\hat{x})\|_2-1)^2]$ is used to enforce the Lipschitz constraint~\cite{gulrajani2017improved}, in which $\hat{x}=\mu x +(1-\mu)G(z,a)$ with $\mu \sim U(0,1)$. As suggested in~\cite{gulrajani2017improved}, we fix $\beta=10$ in this paper.

\begin{figure}[t]
\begin{center}
\includegraphics[width=0.68\linewidth]{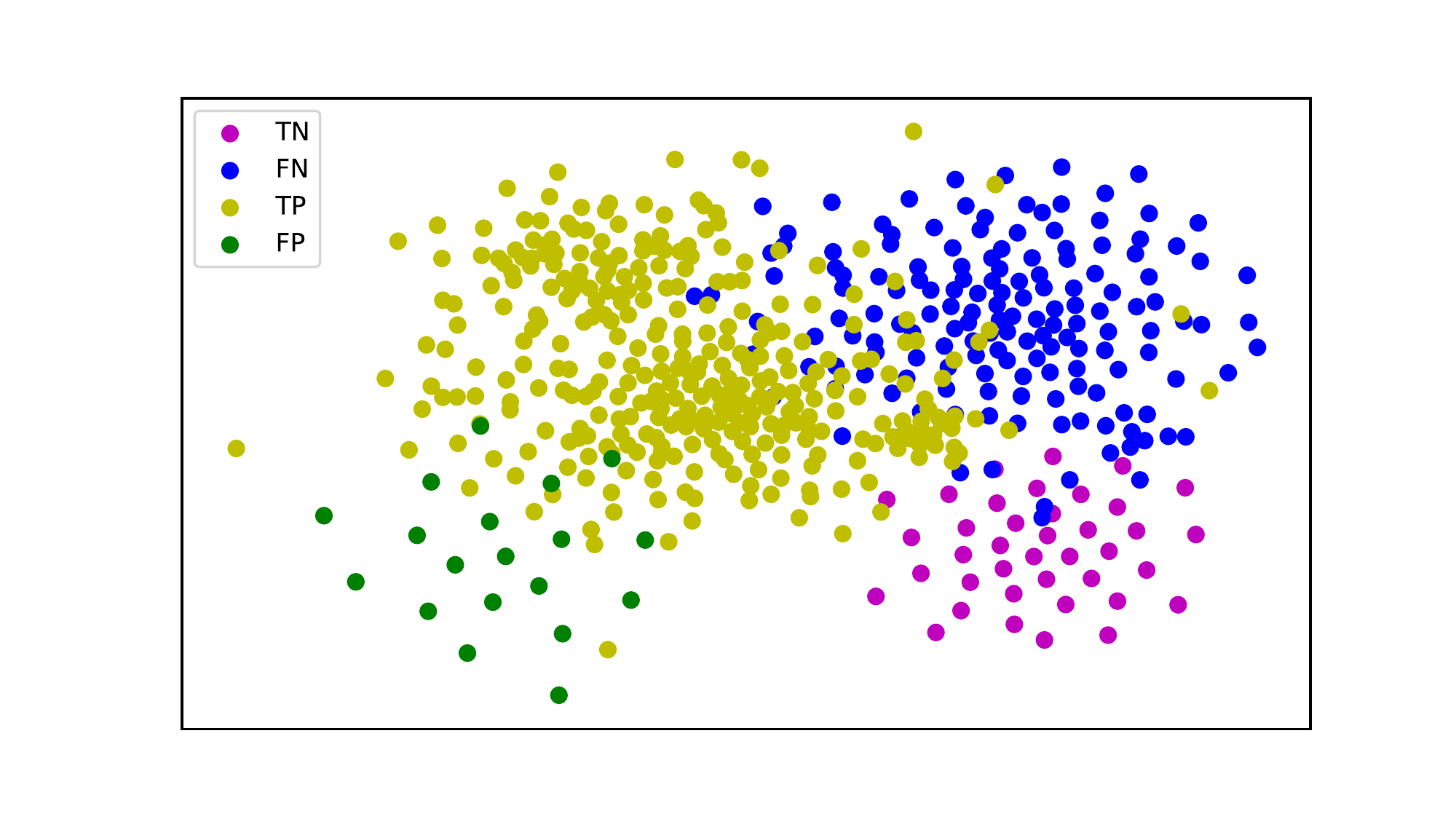}
\end{center}
\vspace{-10pt}
\caption{\small Feature confusion visualization. It can be seen that many unseen instances are misclassified into seen categories (FN). FN holds a high proportion of the FALSE results. It means that unseen categories are much more likely to be misclassified into seen categories than seen categories misclassified into unseen ones.}
\vspace{-10pt} 
\label{fig:confusion}
\end{figure} 

\subsection{The Boundary Loss}
At first, we show the feature confusion issue with both illustrative and quantitative evidences. Fig.~\ref{fig:confusion} shows a visualization of GZSL classification results on CUB, where the synthesized features are generated by state-of-the-art method f-CLSWGAN~\cite{xian2018feature}. It can be seen that too many synthesized unseen features are misclassified into the seen categories. The reason is that the GAN generator is trained on the seen categories, thus, the synthesized features for unseen categories are prone to the seen references. The generated features are confusing. This phenomenon can also be reflected by quantitative results. Taking another state-of-the-art GAZSL~\cite{zhu2018generative} as an example, the accuracy of unseen samples on CUB is 55.8\% when only unseen instances are tested. However, this accuracy drops to 23.9\% when seen categories are also involved in the testing set. It is not hard to speculate that the synthesized unseen features have fairly discriminative information among the unseen categories. However, such discriminative information would be drowned by the feature confusion when seen categories are involved in.

It is no doubt that the feature confusion does exist in generalized zero-shot learning and it undermines the performance of generative models. In this paper, we propose to explicitly handle this problem by introducing a boundary loss which maximizes the decision boundary between seen and unseen categories. To this end, an ideal solution is to pre-train a supervised classifier on seen and unseen instances. However, we do not have any real unseen visual features at the training stage. Inspired by the GAN discriminator, we introduce an additional discriminator $D^\prime$ which distinguishes whether a sample feature is from the seen categories or the unseen categories. Since the discriminator $D'$ has two possible outputs: seen or unseen, we may define its loss as a binary cross-entropy:
\begin{equation}
\label{eq:binary}
  \begin{array}{l} 
  \mathcal{L}_{D'}=  \mathbb{E}[log D'(x_s)]+\mathbb{E}[log(1- D'(G(z,a_u)))], 
  \end{array} 
\end{equation}
where $x_s$ is the real seen sample and $G(z,a_u)$ is the synthesized unseen sample, $G$ is the generator of discriminative WGAN as introduced in Eq.~\eqref{eq:dis}. Instead of `fake'/`real' in classical GAN discriminator, the ground truth supervised information for $D'$ is 'seen'/'unseen'.

It is worth noting that we did not directly train $G$ to minimize $log(1- D'(G(z,a_u)))$ in Eq.~\eqref{eq:binary} since it stands in opposition to our goal. Instead, we only leverage the synthesized features from $G$. By optimizing Eq.~\eqref{eq:binary}, the discriminator $D'$ is able to tell the differences between seen and unseen categories. However, if the generated features are confusing, i.e., $G(z,a_u)$ is similar to $x_s$, the output of the discriminator $D'$ would be ambiguous. Although the final result is binary, either seen or unseen (represented by 1 and 0), the decision is made based on probability. For instance, if the output of the discriminator $D'$ is $[0.9,0.1]$, we would say the sample is from the seen categories since the probability on seen categories (0.9) is significantly higher than unseen categories (0.1). However, when the features are confusing, an ambiguous output like $[0.5,0.5]$ would appear. To minimize the feature confusion, we want a crisp answer of ``seen or unseen'' rather than an ambiguous one. In other words, the probabilities of seen vs. unseen are NOT expected as 0.5 vs. 0.5. Thus, we introduce the boundary loss to prevent such a situation. Formally, we calculate the cross-entropy between the output of $D'$ and the probability [0.5, 0.5]:
 \begin{equation}
\label{eq:fuzzy}
  \begin{array}{l} 
  \mathcal{L}_{boundary}=  -\sum_{k=\{seen, unseen\}} \frac{1}{2} log~D'(\hat{x})_{y=k}
  \end{array} 
\end{equation}  
where $k$ denotes the possible categories, i.e., seen and unseen, $D'(\hat{x})_{y=k}$ is the output of the discriminator $D'$ to predict the probability of seen and unseen for a synthesized feature $\hat{x}$. It is worth noting that minimizing $\mathcal{L}_{boundary}$ tends to encourage feature confusion. Therefore, we maximize it on $G$ and thus inspire $G$ to generate unseen features without confusion with seen ones.

\subsection{The Multi-modal Cycle-loss}
It is well known that the mode collapse is a common issue in GANs. Since ZSL handles unseen/novel instances, the generalization ability of the model is highly critical. With limited diversity, one cannot recognize novel samples which are from complex reality. Recent work~\cite{isola2017image} has verified that an image-to-image cycle-loss is able to mitigate the mode collapse problem. However, we do not have any real images from unseen categories. The only thing we know about unseen categories are their semantic descriptions. Thus, we add a multi-modal cycle-loss to guarantee that the synthesized features can be translated back to their corresponding semantic embeddings. Formally, we define the multi-modal cycle-loss as:
\begin{equation}
\label{eq:cyc}
  \begin{array}{c}
 \mathcal{L}_{cycle}= \mathbb{E}[\|a-\phi(G(z,a))\|_2^2],
  \end{array} 
\end{equation} 
where $\phi(\cdot)$ is a nonlinear mapping. In this paper, we implement it by two FC layers with Rectified Linear Unit (ReLU). 


It is also worth noting that our multi-modal consistent loss is calculated on both synthesized seen features and unseen features. Thus, it is able to further alleviate the feature confusion issue. Specifically, although the GAN generator is mainly trained on seen categories, the multi-modal cycle-consistency involves the semantic embeddings of the unseen categories. By forcing generated seen features pair to seen semantic embeddings and forcing synthesized unseen features to corresponding unseen semantic embeddings, it indeed separates the seen features and unseen features from the semantic conditions (considering that the seen semantic embeddings are different from unseen ones).

\subsection{Overall Objective Function}
The basis of our model is a WGAN which synthesizes virtual features from Gaussian noises conditioned by class embeddings. On this basis, we further introduce a boundary loss to address the feature confusion issue and a multi-modal cycle-loss to handle the mode collapse issue. By fully considering Eq.~\eqref{eq:dis}, Eq.~\eqref{eq:fuzzy} and Eq.~\eqref{eq:cyc}, we have the overall objective for our model:
 \begin{equation}
\label{eq:overall}
  \begin{array}{l} 
  \min\limits_{G}\max\limits_{D,D'}\mathcal{L} = \mathcal{L}_{WGAN}-\eta_1 \mathcal{L}_{boundary}+\eta_2 \mathcal{L}_{cycle},
  \end{array} 
\end{equation}  
where $\eta_1>0$ and $\eta_2>0$ are two penalty parameters.

\subsection{Zero-shot Classification}
By optimizing Eq.~\eqref{eq:overall}, we can train a GAN generator $G$ which is able to synthesize virtual visual features for unseen categories. Then, we use the synthesized features to train a classifier, e.g., softmax, to recognize the real unseen instances. In other words, the zero-shot learning is converted into a supervised classification problem. It is worth noting that such a classification is performed in the visual space. Recall that we trained a nonlinear mapping $\phi(\cdot)$ to calculate the multi-modal cycle-consistent loss, we can also use $\phi(\cdot)$ to map all the visual features into the semantic space and get a classification result. Thus, in our model, we take full advantage of both the semantic-to-visual results $f_v$ and visual-to-semantic results $f_s$ and get the final classification scores:
\begin{equation}
\label{eq:classification}
  \begin{array}{l} 
  f = f_v + \omega f_s,
  \end{array} 
\end{equation}  
where $\omega>0$ is a weight parameter used to balance the two parts. 

For GZSL, the main steps are the same as ZSL. The only difference lies in the final classification process. Specifically, the testing data of GZSL are from both the seen and unseen categories. Thus, we need to train a classifier on both real seen features and synthesized unseen features. 

\subsection{Quantify Feature Confusion}
To the best of our knowledge, there is no previous work which explicitly addresses the feature confusion issue in generative zero-shot learning. Thus, we propose a metric to quantify the feature confusion in this paper. Since the sample numbers are not fixed, a metric which is independent of the number of samples is expected. Instead of concentrating on specific samples, we propose to quantify the feature confusion on class prototypes. Specifically, a class prototype is defined as the mean vector of the embedded support points belonging to its class:
\begin{equation}
\label{eq:prototype}
  \begin{array}{l} 
  c_k=\frac{1}{n_k} \sum_{i=1}^{n_k} x_i^k,
  \end{array} 
\end{equation}  
where $c_k$ indicates the prototype of $k$-th category, $x_i^k$ is the $i$-th sample of $k$-th class and $n_k$ is the sample number of $k$-th class.

Then, we calculate the prototypes of each seen classes and use them as the referring pivots to measure the feature confusion. Formally, we define the feature confusion score (FCS) as:
\begin{equation}
\label{eq:prototype}
  \begin{array}{l} 
  \mathrm{FCS}={\frac{1}{n_u} \sum\limits_{i=1}^{n_u}\min\limits_{j\in[1,k]} \frac{1}{\|\hat{x}_i-c_j\|_2^2}},
  \end{array} 
\end{equation}  
where $n_u$ is the number of synthesized unseen feature, $\hat{x}_i$ is the $i$-th synthesized unseen features, $k$ is the number of seen categories.  

It can be seen that FCS calculates the average reciprocal of distance between each synthesized instances and its nearest seen prototype. Specifically, the more separable between seen and unseen categories, the smaller FCS. In the experiments, we show that our model can significantly reduce the FCS by introducing the boundary loss and multimodal cycle-consistent loss.

\section{Experiments}
In this section, we evaluate our method AFC-GAN on five widely used datasets. The method is tested with both ZSL and GZSL settings. Our codes can be found at {\it github.com/lijin118/AFC-GAN.}

\subsection{Datasets Introduction}
APascal-aYahoo ({\bf aPaY}) contains 32 categories from both PASCAL VOC 2008 dataset and Yahoo image search engine. The total sample number of aPaY is 15,339. An additional 64-dimensional attribute vector is annotated for each category. Animals with Attributes ({\bf AwA})~\cite{lampert2009learning} consists of 30,475 images of $50$ animals classes. The animals classes are aligned with Osherson's classical class/attribute matrix, thereby providing 85 numeric attribute values for each class. Caltech-UCSD Birds-200-2011 ({\bf CUB})~\cite{wah2011caltech} contains 11,788 images of 200 bird species. A vocabulary of 28 attribute groupings and 312 binary attributes were associated with the dataset based on an online tool for bird species identification. Oxford Flowers ({\bf FLO})~\cite{Nilsback08} dataset consists of 8,189 images which comes from 102 flower categories. Each class consists of 40$\sim$258 images. For this dataset, we use the same semantic descriptions provided by Reed et al.~\cite{reed2016learning}. SUN attributes ({\bf SUN})~\cite{patterson2012sun} is a large-scale scene attribute dataset, which spans 717 categories and 14,340 images in total. Each category includes 102 attribute labels. 

For clarity, we report the dataset statistics and zero-shot split settings in Table~\ref{tab:dataset}. The zero-shot splits of aPaY, AwA, CUB and SUN are same with previous work~\cite{xian2017zero} and the splits of FLO is same with~\cite{reed2016learning}. For the real CNN features, we follow previous work~\cite{xian2018feature} to extract 2048-dimensional features from ResNet-101~\cite{he2016deep} which is pre-trained on ImageNet. For the semantic descriptions, we use the default attributes included in the datasets. Specifically, since FLO did not provide attributes with the dataset, we use the 1024-dimensional RNN descriptions via the model of~\cite{reed2016learning}. For fair comparisons, all of our experimental settings are kept the same as reported in~\cite{xian2018feature}.

\renewcommand\arraystretch{1}
\begin{table}[t!p]
\centering
\caption{Dataset statistics. The (number) in \# Seen Classes indicates the number of seen classes used for test in the generalized zero-shot learning. }
\vspace{-8pt}
\label{tab:dataset}
\begin{tabular}{lccccc}
\toprule
Dataset & aPaY & AwA & CUB & FLO & SUN  \\
\midrule
\# Samples & 15,339 & 30,475 & 11,788 & 8,189 & 14,340   \\
\# Attributes & 64 & 85 & 312 & 1,024 & 102  \\
\# Seen Classes & 20 (5) & 40 (13) & 150 (50) & 82 (20) & 645 (65)  \\     
\# Unseen Classes\!\!\!\!\! & 12 & 10 & 50 & 20 & 72  \\
\bottomrule
\end{tabular}
\vspace{-10pt}
\end{table}

\subsection{Implementation and Compared Methods}
In our model, the GAN is implemented via multilayer perceptron with Rectified Linear Unit (ReLU) activation. Specifically, the generator $G$ contains a fully connected layer with 4,096 hidden units. The noise $z$ is sampled from Gaussian distributions and then conditioned on the semantic description $a$ and then severed as the inputs of $G$. An additional ReLU layer is deployed as the output layer of $G$ which outputs the synthesized virtual features. The discriminator $D$ takes the real features and the synthesized virtual features from $G$ and processes them via an FC layer, a Leaky ReLU layer, an FC layer and a ReLU layer. The discriminator has two branches for output. One is used to tell synthesized from real and the other is a standard $n$-ways classifier to predict the correct category of each sample. The additional discriminator $D'$ is implemented by an FC layer with ReLU. The nonlinear deep mapping from visual space to semantic space is implemented by two FC layers with ReLU. 

The compared methods are representative ones published in the fast few years and the state-of-the-art ones reported very recently. Specifically, we compare our approach with: DAP~\cite{lampert2014attribute}, CONSE~\cite{norouzi2013zero}, SSE~\cite{zhang2015zero}, DeViSE~\cite{frome2013devise}, SJE~\cite{akata2015evaluation}, ESZSL~\cite{romera2015embarrassingly}, ALE~\cite{akata2016label}, SYNC~\cite{changpinyo2016synthesized}, SAE~\cite{kodirov2017semantic}, GAZSL~\cite{zhu2018generative} and f-CLSWGAN~\cite{xian2018feature}. 

Following previous work~\cite{xian2018feature,zhu2018generative}, we report the average per-class top-1 accuracy for each of the evaluated methods. Specifically, for classical zero-shot learning, we report the top-1 accuracy of unseen samples by only searching the unseen label space. However, for the generalized zero-shot learning, we report the accuracy on both seen classes and unseen classes with the same settings in~\cite{xian2017zero}. Some of the results reported in this paper are also cited from the survey paper~\cite{xian2017zero}. Our codes and corresponding datasets will be publicly available on Github for the convenience of readers.

\begin{table}[t!p]
\centering
\caption{The results (top-1 accuracy \%) of ZSL on different datasets. The best results are highlighted in bold.} 
\vspace{-8pt}
\label{tab:zsl}
\begin{tabular}{lccccc}
\toprule
Methods & ~~~aPaY~~~ & ~~~AwA~~~ & ~~~CUB~~~ & ~~~FLO~~~ & ~~~SUN~~~  \\
\midrule
DAP~\cite{lampert2014attribute} & 33.8 & 44.1 & 40.0 & - & 39.9   \\
CONSE~\cite{norouzi2013zero} & 26.9 & 45.6 & 34.3 & - & 38.8  \\
SSE~\cite{zhang2015zero} & 34.0 & 60.1 & 43.9 & - & 51.5  \\     
DeViSE~\cite{frome2013devise} & 39.8 & 54.2 & 52.0 & 45.9 & 56.5  \\

SJE~\cite{akata2015evaluation} & 32.9 & 65.6 & 53.9 & 53.4 & 53.7   \\
ESZSL~\cite{romera2015embarrassingly} & 38.3 & 58.2 & 53.9 & 51.0 & 54.5  \\
ALE~\cite{akata2016label} & 39.7 & 59.9 & 54.9 & 48.5 & 58.1  \\     
SYNC~\cite{changpinyo2016synthesized} & 23.9 & 54.0 & 55.6 & - & 56.3  \\

SAE~\cite{kodirov2017semantic} & 8.3 & 53.0 & 33.3 & - & 40.3   \\
DEM~\cite{zhang2017learning} & 35.0 & 68.4 & 51.7 & - & { 61.9}  \\
GAZSL~\cite{zhu2018generative} & 41.1 & 68.2 & 55.8 & 60.5 & 61.3  \\     
f-CLSWGAN~\cite{xian2018feature} & 40.5 & 68.2 & 57.3 & 67.2 & 60.8  \\
\midrule
AFC-GAN~[Ours]  & {\bf 45.5} & {\bf 69.1} & {\bf 62.9} & {\bf 70.5} &  {\bf 63.3}  \\
\bottomrule
\end{tabular}
\vspace{-10pt}
\end{table}

\begin{table*}[t]
\centering
\caption{The results (top-1 accuracy \%) of GZSL. The Mean in this table is the harmonic mean of seen and unseen samples, i.e., Mean=(2*Unseen*Seen)/(Unseen+Seen). The best results are highlighted with bold numbers. } 
\vspace{-8pt}
\label{tab:gzsl}
\small \begin{tabular}{l|ccc|ccc|ccc|ccc|ccc}
\toprule
\multirow{2}{*}{Methods} & \multicolumn{3}{c|}{aPaY} & \multicolumn{3}{c|}{AwA} & \multicolumn{3}{c|}{CUB} & \multicolumn{3}{c|}{FLO} & \multicolumn{3}{c}{SUN}  \\
\cline{2-16}
 & Unseen & Seen & Mean & Unseen & Seen & Mean & Unseen & Seen & Mean & Unseen & Seen & Mean & Unseen & Seen & Mean \\
 \midrule 
DAP~\cite{lampert2014attribute} & 4.8 & 78.3 & 9.0 & 0.0 & 88.7 & 0.0 & 1.7 & 67.9 & 3.3 & - & - & - & 4.2 & 25.2 & 7.2  \\
CONSE~\cite{norouzi2013zero}  & 0.0 & {\bf 91.2} & 0.0 & 0.4 & 88.6 & 0.8 & 1.6 & 72.2 & 3.1 & - & - & - & 6.8 & 39.9 & 11.6\\
SSE~\cite{zhang2015zero} & 0.2 & 78.9 & 0.4 & 7.0 & 80.5 & 12.9 & 8.5 & 46.9 & 14.4 & - & - & - & 2.1 & 36.4 & 4.0  \\     
DeViSE~\cite{frome2013devise} & 4.9 & 76.9 & 9.2 & 13.4 & 68.7 & 22.4 & 23.8 & 53.0 & 32.8 & 9.9 & 44.2 & 16.2 & 16.9 & 27.4 & 20.9  \\

SJE~\cite{akata2015evaluation} & 3.7 & 55.7 & 6.9 & 11.3 & 74.6 & 19.6 & 23.5 & 59.2 & 33.6 & 13.9 & 47.6 & 21.5 & 14.7 & 30.5 & 19.8   \\
ESZSL~\cite{romera2015embarrassingly} & 2.4 & 70.1 & 4.6 & 6.6 & 75.6 & 12.1 & 2.4 & 70.1 & 4.6 & 11.4 & 56.8 & 19.0 & 11.0 & 27.9 & 15.8  \\
ALE~\cite{akata2016label} & 4.6 & 73.7 & 8.7 & 16.8 & 76.1 & 27.5 & 4.6 & 73.7 & 8.7 & 13.3 & 61.6 & 21.9 &21.8 & 33.1 & 26.3  \\     
SYNC~\cite{changpinyo2016synthesized} & 7.4 & 66.3 & 13.3 & 8.9 & {\bf 87.3} & 16.2 & 7.4 & 66.3 & 13.3 & - & - & - & 7.9 & {\bf 43.3} & 13.4  \\

SAE~\cite{kodirov2017semantic} & 0.4 & {80.9} & 0.9 & 1.8 & 77.1 & 3.5 & 0.4 & {\bf 80.9} & 0.9 & - & - & - & 8.8 & 18.0 & 11.8   \\
GAZSL~\cite{zhu2018generative} & 14.2 & 78.6 & 24.0 & 19.2 & 86.5 & 31.4 & 23.9 & 60.6 & 34.3 & 28.1 & 77.4 & 41.2 & 21.7 & 34.5 & 26.7  \\     
f-CLSWGAN~\cite{xian2018feature} & 32.9& 61.7 & 42.9 & {57.9} & 61.4 & 59.6 & 43.7 & 57.7 & 49.7 & 59.0 & 73.8 & 65.6 & 42.6 & 36.6 & 39.4  \\
\midrule

AFC-GAN~[Ours]  & {\bf 36.5} & 62.6 & {\bf 46.1} & {\bf 58.2} & 66.8 & {\bf 62.2} & {\bf 53.5} & 59.7 & {\bf 56.4} & {\bf 60.2} &  {\bf 80.0} & {\bf 68.7} & {\bf 49.1} & 36.1 & {\bf 41.6}  \\
\bottomrule
\end{tabular}
\end{table*}

\subsection{Zero-shot Learning}
The results of ZSL are reported in Table~\ref{tab:zsl}. Since generative zero-shot learning is a new topic in the community, most of the compared methods are from the embedding group. GAZSL, f-CLSWGAN and our method are three generative methods. From the results, we can see that generative methods work stably better on all the five datasets. Among the three generative methods, our AFC-GAN performs the best on all the evaluations. It is worth noting that GAZSL, f-CLSWGAN and AFC-GAN share the same basic architecture, i.e., WGAN. The outstanding performance of our AFC-GAN proves that our motivation is reasonable and our formulation is effective. 

In summary, we achieve 5.0\%, 0.9\%, 5.6\%, 3.3\% and 2.5\% improvements on aPaY, AwA, CUB, FLO and SUN, respectively, against the previous state-of-the-art. The average improvement over the five datasets is 3.46\%. We would like to highlight that our performance improvement against f-CLSWGAN are owed to the efforts of alleviating feature confusion. Specifically, we introduce the boundary loss and the multi-modal cycle-consistent loss. The specific contribution of each part will be discussed in the ablation study.

\vspace{-5pt}
\subsection{Generalized Zero-shot Learning}
The results of generalized zero-shot learning are reported in Table~\ref{tab:gzsl}. Following previous work~\cite{xian2017zero}, we report the harmonic mean as the main metric. Please note that harmonic mean can avoid the effects of extreme values. For instance, we can see from the results that SAE gets 0.4\% for unseen and 80.9\% for seen on CUB. Although the accuracy of seen is the best, the harmonic mean is only 0.9\% due to the extremely low result on unseen categories. In a nutshell, the harmonic mean is high only when the accuracies on both seen and unseen categories are high. From the results, we can observe that our method achieves the best harmonic mean on all of the evaluations. It indicates that our AFC-GAN is a stable method which can work well for both seen and unseen instances. In terms of numbers, AFC-GAN outperforms state-of-the-art f-CLSWGAN with 3.2\%, 2.6\%, 6.7\%, 3.1\% and 2.2\% improvements on aPaY, AwA, CUB, FLO and SUN, respectively. The average improvement over five datasets is 3.56\%. Notably, our method achieves the best on all of the unseen categories. We significantly outperform f-CLSWGAN 9.8\% and 6.5\% on CUB and SUN with regard to unseen categories.

Beyond the results in Table~\ref{tab:gzsl} alone, it is more interesting to compare the results in Table~\ref{tab:zsl} and Table~\ref{tab:gzsl}. Please notice that the results in Table~\ref{tab:zsl} are the results on unseen categories under the settings of conventional ZSL. The only difference between the ZSL and GZSL is that samples from both seen and unseen classes are tested in GZSL while only unseen classes are tested in ZSL. Thus, we can see the effects of involving seen classes in testing by comparing the results in Table~\ref{tab:zsl} and the results of unseen in Table~\ref{tab:gzsl}. Let us compare Table~\ref{tab:zsl} and Table~\ref{tab:gzsl} from top to bottom. We take the results on CUB as examples. It is somewhat out of the blue that the results drops dramatically. For instance, DAP drops from 40.0\% to 1.7\%, CONSE drops from 34.3\% to 1.6\%, ..., GAZSL drops from 68.2\% to 23.9\% and f-CLSWGAN drops from 68.2\% to 43.7\%. It is also worth noting that the results of almost all methods on seen categories are significantly better than the results on unseen categories. In fact, the two observations, performance-dropping-on-unseen-categories and the performance-better-on-seen-categories, have exactly the same reason: unseen instances are misclassified into seen categories. 

Although generative ZSL methods are able to alleviate the issue by generating unseen features, e.g., f-CLSWGAN significantly improves the results on unseen categories while maintaining a good performance on seen categories, the performance dropping of generative ZSL methods is still considerable. As we have discussed in the previous contexts, the performance dropping of generative ZSL methods is caused by feature confusion. In our model, we introduce two strategies to alleviate the feature confusion. The first one is the boundary loss which maximizes the decision boundary of seen and unseen categories. The second one is the multi-modal cycle-consistent loss. Comparing the results of f-CLSWGAN and our AFC-GAN, we can see that the two strategies significantly improve the performance on unseen categories.

\subsection{Model Analysis}
{\bf Feature Confusion Score.} In this paper, we proposed a metric named feature confusion score (FCS) to quantify the feature confusion. FCS evaluates the distance between synthesized samples and seen prototypes. As a result, it is able to show the degree of feature confusion. For fair comparisons, we randomly select $1,000$ generated samples from f-CLSWGAN and our method then report corresponding FCS in Table~\ref{tab:fcs}. It can be seen that our method can significantly reduce the FCS on various categories, which verifies that our method is effective to alleviate the feature confusion issue.

\renewcommand\arraystretch{1.0}
\begin{table}[t!p]
\centering
\caption{Feature confusion score (FCS), which tests the confusion degree of generated features. The smaller the better.}
\vspace{-8pt}
\label{tab:fcs}
\begin{tabular}{lccccc}
\toprule
Dataset & ~~~aPaY~~~ & ~~~~AwA~~~~ & ~~~~CUB~~~~ & ~~~~FLO~~~~ & ~~~~SUN~~~~  \\
\midrule
f-CLSWGAN & 0.29 & 0.35 & 0.37 & 0.29 & 0.34  \\     
AFC-GAN~[Ours] & 0.26 & 0.31 & 0.32 & 0.25 & 0.32  \\
\bottomrule
\end{tabular}
\vspace{-10pt}
\end{table}

\begin{figure*}[t!h]
\begin{center}
\subfigure[Boundary loss ($\eta_1$)]{
\includegraphics[width=0.188\linewidth]{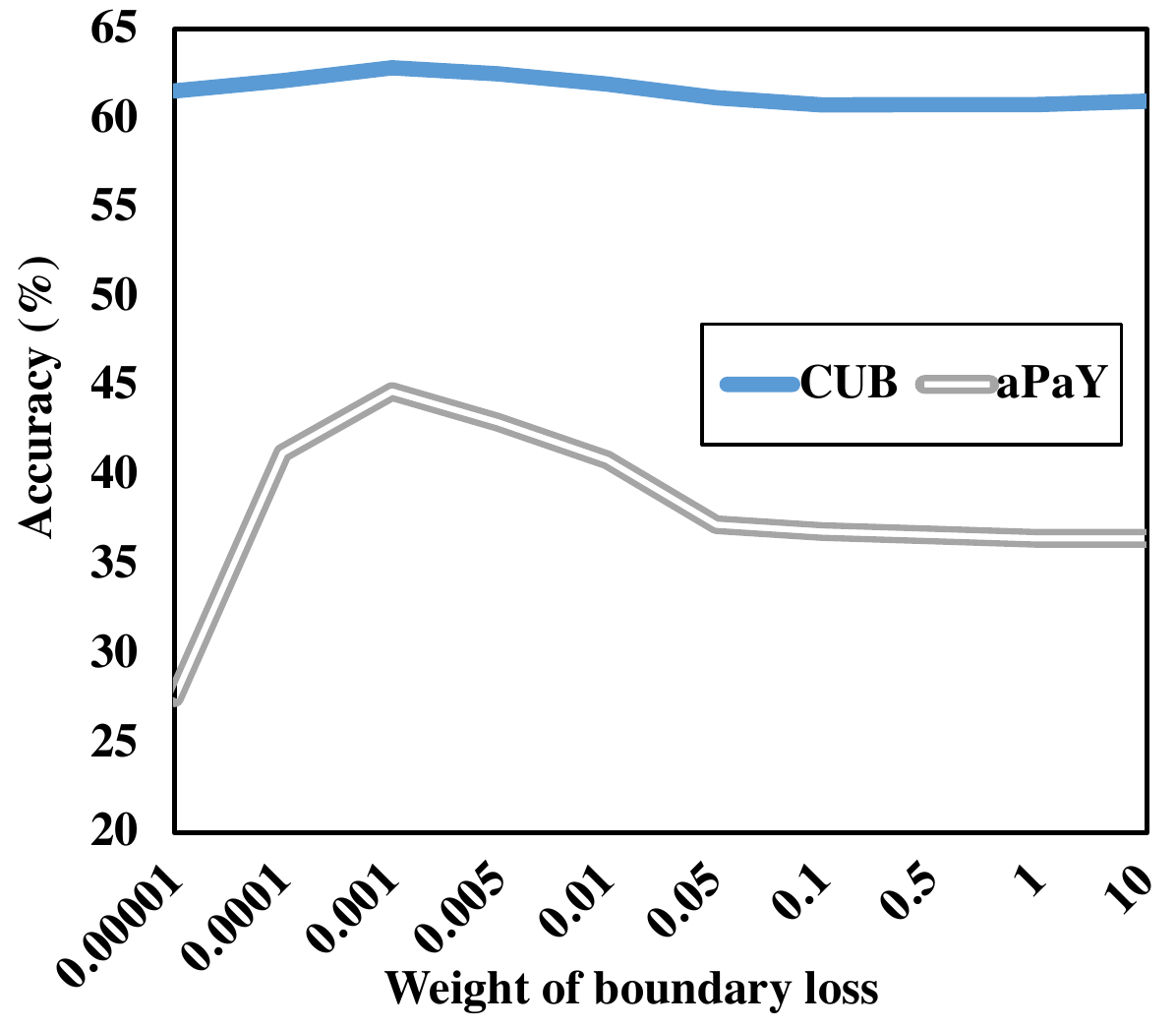}
}
\subfigure[Cycle-loss ($\eta_2$)]{
\includegraphics[width=0.183\linewidth]{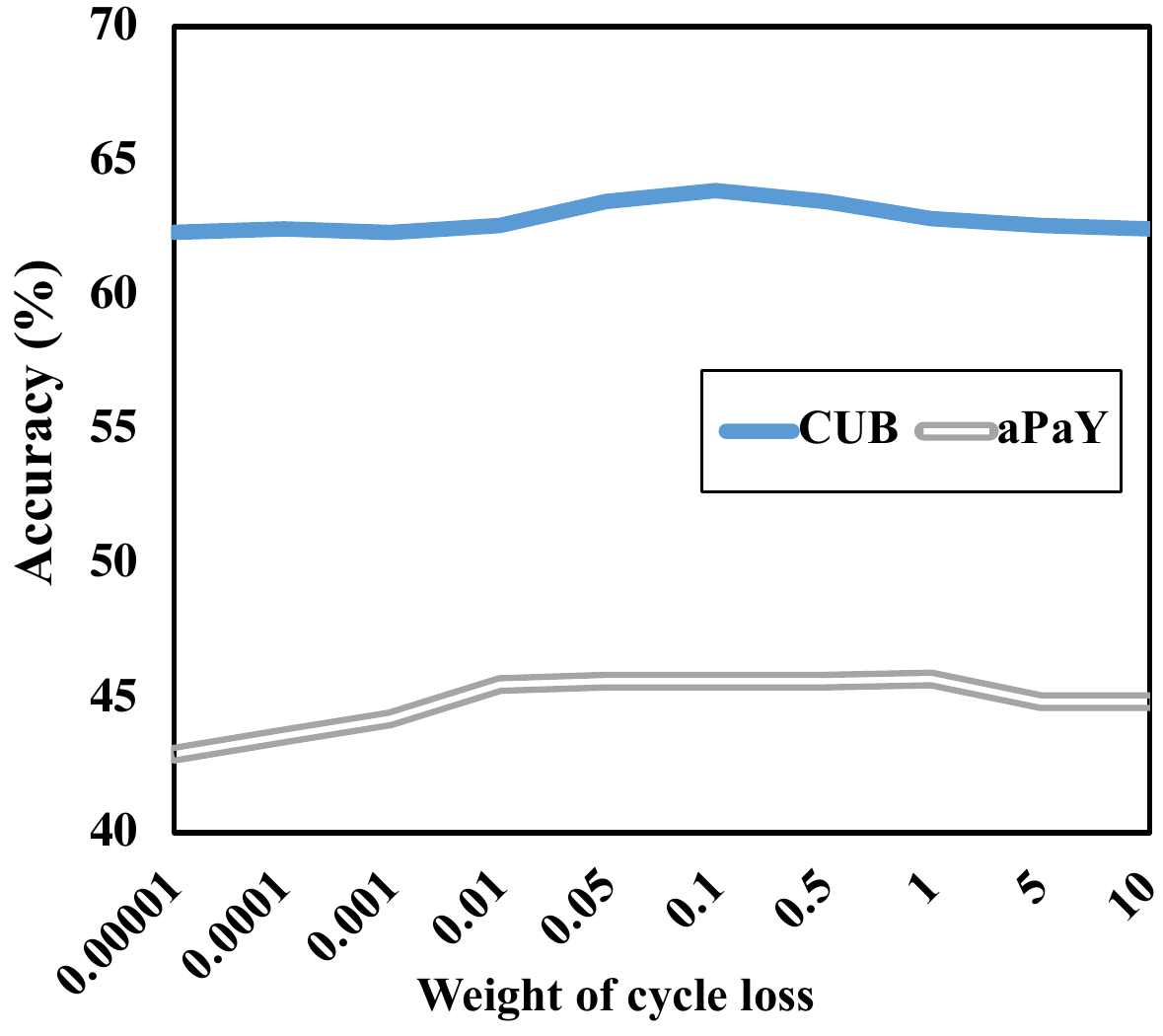}
}
\subfigure[Classifier weight ($\omega$)]{
\includegraphics[width=0.188\linewidth]{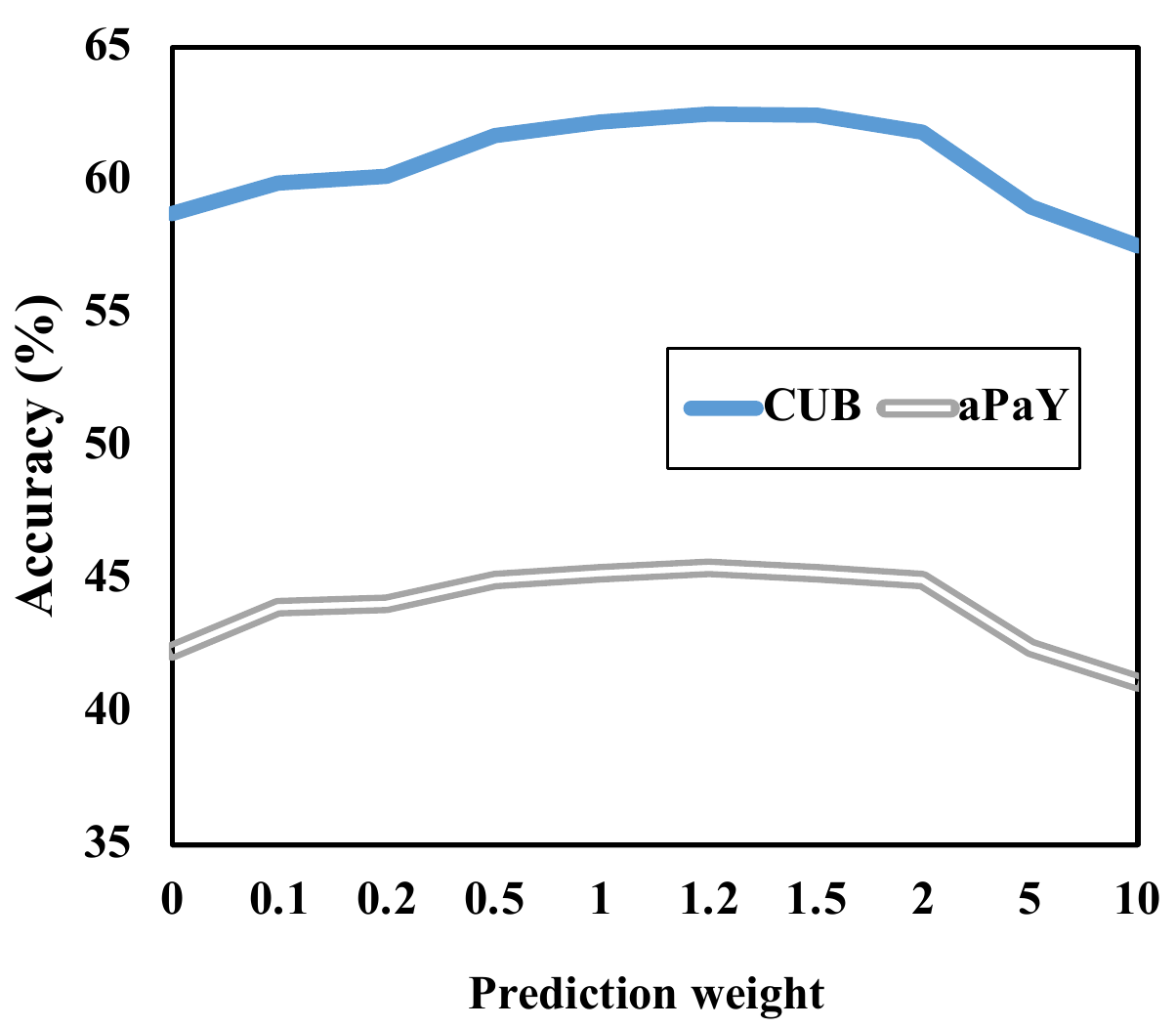}
}
\subfigure[Errors of ZSL]{
\includegraphics[width=0.19\linewidth]{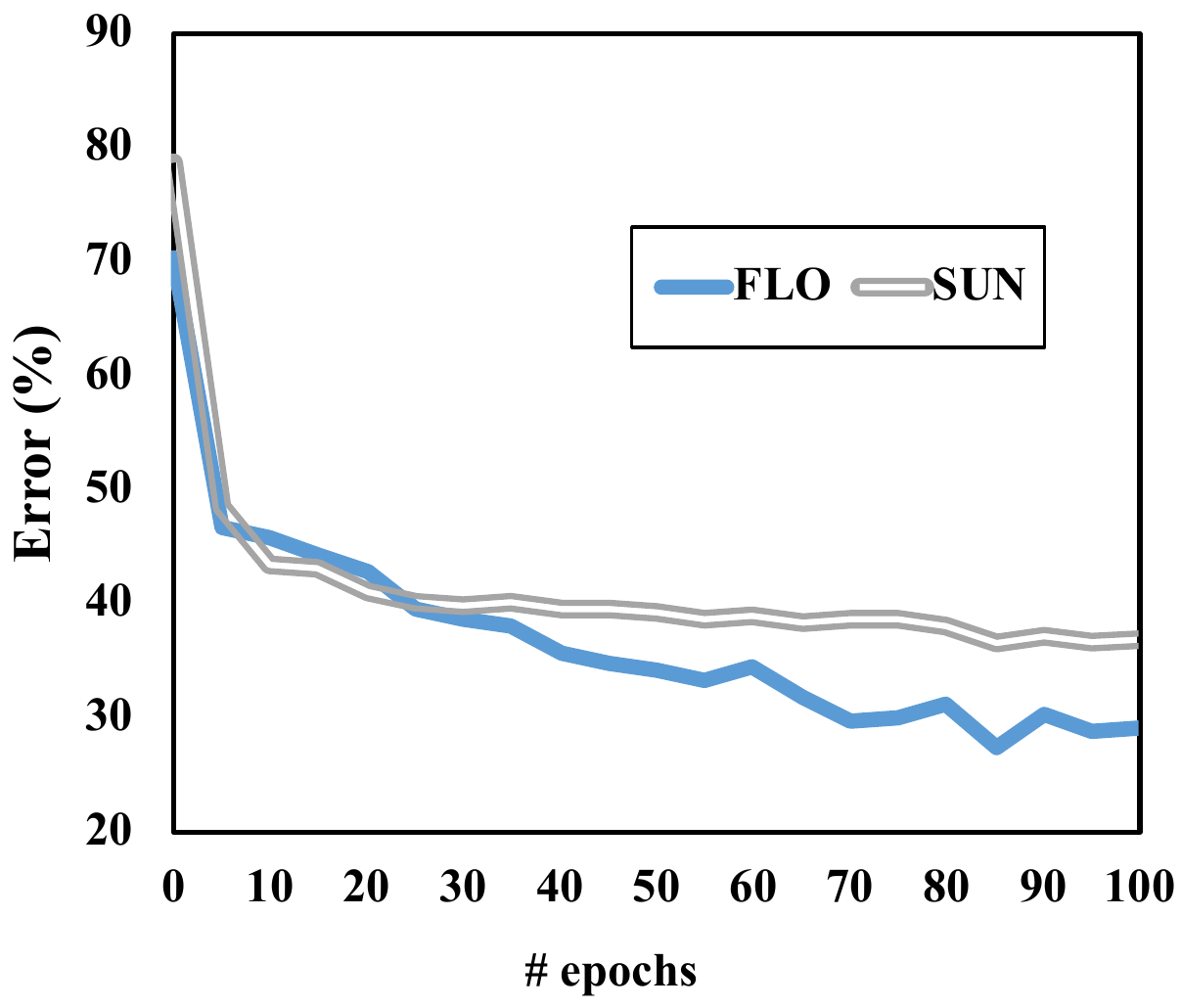}
}
\subfigure[Errors of GZSL]{
\includegraphics[width=0.19\linewidth]{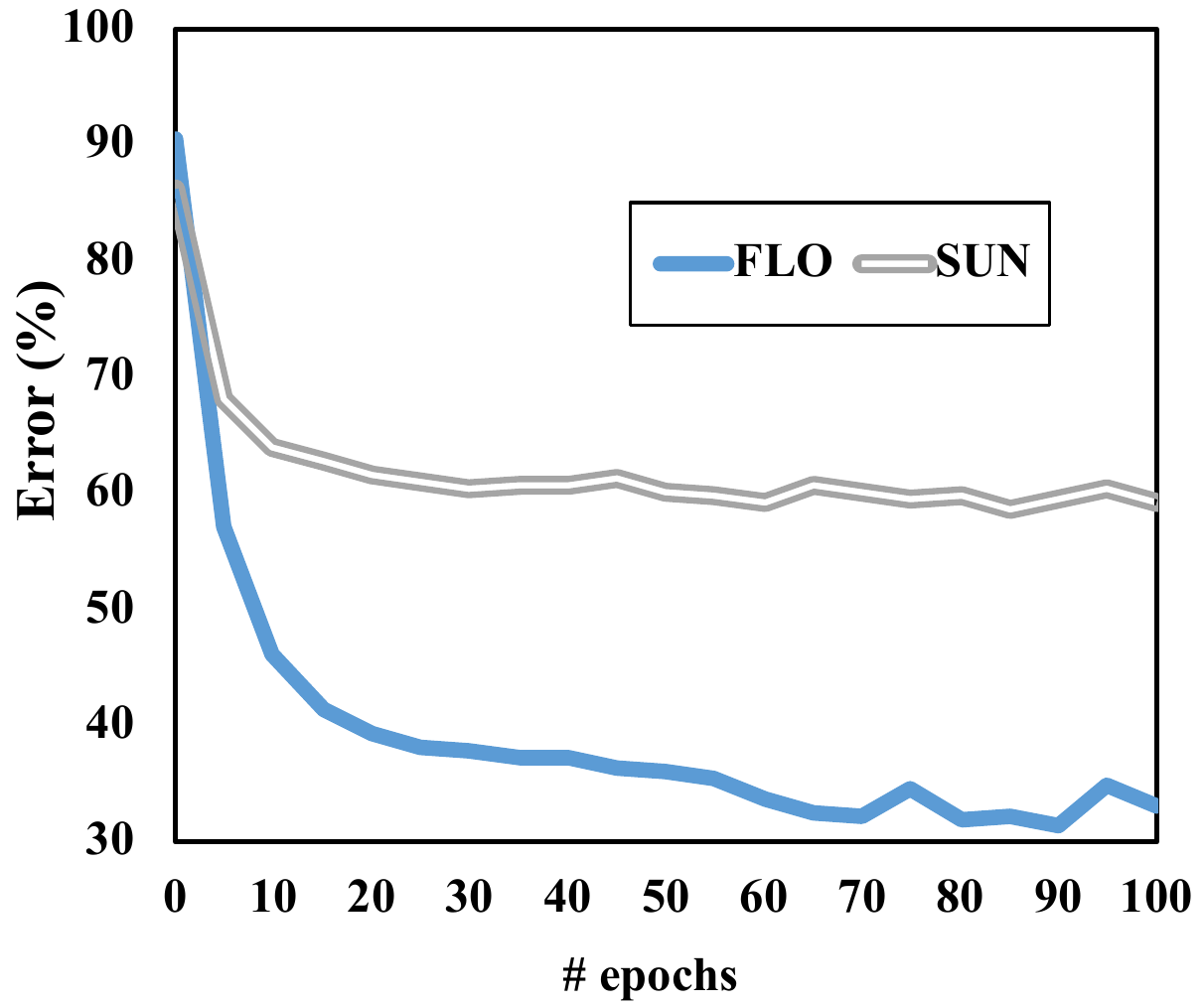}
}
\end{center}
\vspace{-16pt}
\caption{Parameter sensitivity (a-c) and training stability (d-e). Limited by space, two datasets per figure are used for tests. }
\label{fig:para}
\vspace{-10pt}
\end{figure*}

\begin{figure*}[t!h]
\begin{center}
\includegraphics[width=0.98\linewidth]{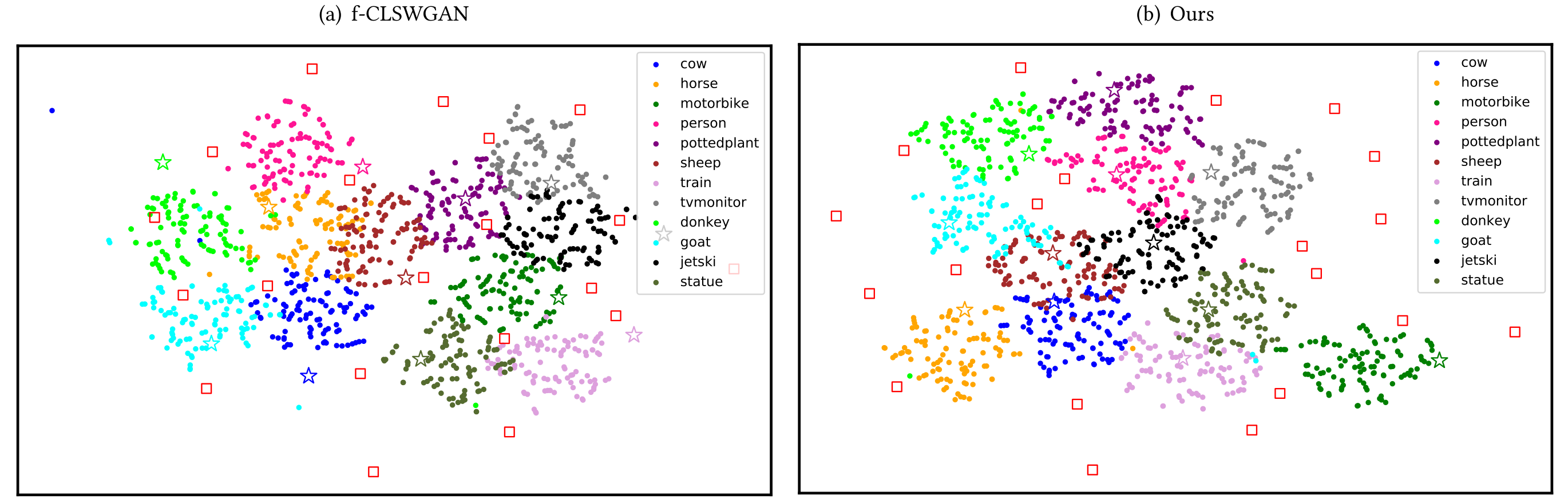}
\vspace{-8pt}
\caption{Visualization of synthesized features on aPaY. Different colors denote different categories. The solid dots, hollow squares and hollow stars represent synthesized features, the prototypes of real seen categories and the prototypes of real unseen categories, respectively. It can be seen that our method can alleviate the feature confusion issue, i.e., the synthesized features of our method are closer to the corresponding prototypes of true category prototypes.}
\label{fig:tsne}
\vspace{-5pt}
\end{center}
\end{figure*}

{\bf Parameters.} The parameters of our method are tuned by importance weighted cross validation. For fair comparisons, we keep $\lambda$ as the same value in f-CLSWGAN~\cite{xian2018feature}. We mainly tune $\eta_1$, $\eta_2$ and $\omega$ for our method. The parameter sensitivity of them are shown in Fig.~\ref{fig:para}(a)-Fig.~\ref{fig:para}(c). It can be seen that our method can work stably with different parameters. A general suggestion is that the these parameters are better chosen from (0,1].

{\bf Training Stability.} Since GAN-based methods are hard for training, we report the training process of our method in Fig.~\ref{fig:para}(d) and Fig.~\ref{fig:para}(e). Specifically, Fig.~\ref{fig:para}(d) reports the evaluation errors of ZSL with different epochs, and Fig.~\ref{fig:para}(e) shows the training trend of GZSL with different epochs. From the results, we can see that our method enjoys fast convergence. It generally can achieve stable performance within 100 epochs under both ZSL and GZSL settings.

{\bf Visualization.} We visualize the synthesized features on aPaY dataset by t-SNE~\cite{maaten2008visualizing} and report the results in Fig.~\ref{fig:tsne}. Specifically, we generate $50$ samples for each unseen class and visualize the generated features. In Fig.~\ref{fig:tsne}, we also show the prototypes of seen classes and unseen classes in hollow squares and hollow stars, respectively, it can be seen that the generated features of our method have more clear distance with the seen prototypes and are prone to true unseen prototypes, which verifies that our method is effective for alleviating feature confusion.

{\bf Ablation Study.} We report the ablation study of our method under both ZSL and GZSL settings in Table~\ref{tab:ablation}. In the reported results, S1 is the very basic model. S2 and S3 are proposed to alleviate the feature confusion issue. S4 reflects the results of Eq.~\eqref{eq:classification}. It refines the classification results of visual space and semantic space. From the results, it is safe to draw the conclusion that our proposals significantly improve the performance of baseline S1. Comparing the results of ZSL and GZSL, we can find that the improvements on GZSL are more remarkable than that on ZSL, which further verifies the effectiveness of our method on handling feature confusion issue in generative zero-shot learning. Our method is able to promote the performance on unseen categories and, at the same time, maintain the accuracy on seen categories. In fact, the results in Table~\ref{tab:gzsl} show that our method achieves the best on all of the unseen categories.

{\bf Effectiveness of The Boundary Loss.} As a major contribution of this paper, the boundary loss makes our method stands out from previous approaches, especially from f-CLSWGAN. After reporting the experimental results in Table~\ref{tab:gzsl}, Table~\ref{tab:ablation} and Fig.~\ref{fig:tsne}, we can analyze the effectiveness of the boundary loss in both quantitative and qualitative manners. We argue that the boundary loss is effective to alleviate the feature confusion issue, which indicates that many unseen instances are misclassified into seen categories. From the results in Table~\ref{tab:gzsl}, we can clearly see that our method improves the performance on unseen categories, e.g., the accuracies of unseen categories on CUB and SUN are improved by 9.8\% and 6.5\%, respectively, which demonstrates the significance of solving the feature confusion. Furthermore, from the ablation study in Table~\ref{tab:ablation}, we can quantitatively evaluate the contribution of the boundary loss by comparing the results of S1 and S2, which further verifies the effectiveness of the boundary loss. For instance, the overall performance of S2 (with the boundary loss) on CUB is 3.5\% better than S1 (without the boundary loss). The results in Table~\ref{tab:fcs} also demonstrate that the boundary loss is useful to reduce the feature confusion score. At last, for a better understanding, Fig.~\ref{fig:tsne} vividly shows that the boundary loss is effective for discriminating unseen samples from seen categories. In other words, it is effective for alleviating feature confusion.


\begin{table}[t!p]
\centering
\caption{The results of ablation study on CUB and FLO. The result of GZSL is reported as the harmonic mean.}
\vspace{-8pt}
\label{tab:ablation}
\begin{tabular}{clcc}
\toprule
\multicolumn{2}{l}{Settings} & CUB & FLO  \\
\toprule
\multirow{4}{*}{ZSL} & S1: Discriminative Conditional WGAN\hspace{3pt}  & 57.3 & 67.2   \\
& S2: S1 + boundary loss  & 60.2 & 68.5  \\     
& S3: S2 + cycle-consistent loss  & 61.5 & 69.3  \\
& S4: S3 + classification refining  & 62.9 & 70.5  \\
\midrule
\multirow{4}{*}{GZSL}\hspace{3pt} & S1: Discriminative Conditional WGAN  & 49.7 & 65.6   \\
& S2: S1 + boundary loss  & 53.2 & 66.7  \\     
& S3: S2 + cycle-consistent loss  & 54.7 & 67.5  \\
& S4: S3 + classification refining  & 56.4 & 68.7  \\
\bottomrule
\end{tabular}
\vspace{-15pt}
\end{table}


\section{Conclusion and Future Work}
In this paper, we proposed a novel zero-shot learning method named alleviating feature confusion GAN (AFC-GAN) by taking advantage of adversarial generative networks. This work explicitly addresses the feature confusion issue and proposes to handle it by introducing a boundary loss and a multi-modal cycle-consistent loss. In addition, we also proposed a new metric to quantify feature confusion. Experiments on five widely used datasets verify that our method is able to effectively alleviate the feature confusion issue. To the best of our knowledge, this is the first work which explicitly defines and challenges the feature confusion issue in generative zero-shot learning. In our future work, we will work on learning a better classifier, rather than a straightforward binary classifier, to distinguish seen and unseen categories, so that GZSL can be directly converted into two supervised deep learning problems.
%

\section{Acknowledgments}
This work was supported in part by the NSFC under Grant 61806039, 61832001, 61572108 and 61632007, in part by ARC under DP190102353, in part by the National Postdoctoral Program for Innovative Talents under Grant BX201700045, in part by the China Postdoctoral Science Foundation under Grant 2017M623006 and in part by Sichuan Department of Science and Technology under Grant 2019YFG0141 and 2018GZDZX0032.
%

\balance
\bibliographystyle{ACM-Reference-Format}
\bibliography{mm19}

%

\end{document}